\documentclass[10pt,twocolumn,letterpaper]{article}

\usepackage{iccv}              
\usepackage{multirow}
\usepackage{indentfirst}
\usepackage{float}

\definecolor{iccvblue}{rgb}{0.21,0.49,0.74}
\definecolor{first}{HTML}{EDD5C5}
\definecolor{second}{HTML}{B1C2D5}
\usepackage[pagebackref,breaklinks,colorlinks,allcolors=iccvblue]{hyperref}
\newcommand{\figref}[1]{Fig.~\ref{#1}}
\newcommand{\tabref}[1]{Tab.~\ref{#1}}

\newcommand{\secref}[1]{Sec.~\ref{#1}}

\def\paperID{6416} 
\def\confName{ICCV}
\def\confYear{2025}

\title{SDMatte: Grafting Diffusion Models for Interactive Matting}

\author{
  Longfei Huang$^{1,2}$\thanks{Equal contribution. Intern at vivo Mobile Communication Co., Ltd.} ~~ 
  Yu Liang$^2$\footnotemark[1] ~~ Hao Zhang$^2$ ~~ Jinwei Chen$^2$ ~~ Wei Dong$^2$ ~~ \\
  Lunde Chen$^1$ ~~ Wanyu Liu$^1$ ~~ Bo Li$^2$ ~~ 
  Peng-Tao Jiang$^2$\thanks{Peng-Tao Jiang is the corresponding author.} \\
$^1$Shanghai University \quad
$^2$vivo Mobile Communication Co., Ltd. \\
{\tt\small 2946399650fly@shu.edu.cn  \quad \quad pt.jiang@vivo.com }}

\begin{document}
\maketitle
\begin{abstract}
\indent Recent interactive matting methods have shown satisfactory performance in capturing the primary regions of objects, but they fall short in extracting fine-grained details in edge regions.
Diffusion models trained on billions of image-text pairs, demonstrate exceptional capability in modeling highly complex data distributions and synthesizing realistic texture details, while exhibiting robust text-driven interaction capabilities, making them an attractive solution for interactive matting. 
To this end, we propose SDMatte, a diffusion-driven interactive matting model, with three key contributions.
First, we exploit the powerful priors of diffusion models and transform the text-driven interaction capability into visual prompt-driven interaction capability to enable interactive matting.
Second, we integrate coordinate embeddings of visual prompts and opacity embeddings of target objects into U-Net, enhancing SDMatte's sensitivity to spatial position information and opacity information.
Third, we propose a masked self-attention mechanism that enables the model to focus on areas specified by visual prompts, leading to better performance.
Extensive experiments on multiple datasets demonstrate the superior performance of our method, validating its effectiveness in interactive matting.
Our code and model are available at \href{https://github.com/vivoCameraResearch/SDMatte}{https://github.com/vivoCameraResearch/SDMatte}.
\end{abstract}    
\section{Introduction}
\label{sec:intro}
Image matting, as a fundamental task of computer vision, involves estimating a precise alpha matte to separate the foreground from the background and has attracted significant research interest. 
However, because of the unknown nature of the foreground, background, and alpha matte, image matting constitutes a highly ill-posed problem.

To address this problem, DIM~\cite{xu2017deep} first introduces a trimap as an auxiliary input, which explicitly divides the image into three regions: definite foreground, definite background, and unknown region that needs to be predicted. 
Given that the semantic guidance provided by trimaps substantially reduces the difficulty of the image matting task, subsequent studies~\cite{tang2019learning, forte2020f, yao2024vitmatte, hu2025diffusion} have adopted the DIM framework, utilizing trimaps as auxiliary input to predict high-quality alpha mattes.
Although trimaps significantly improve the accuracy of alpha matte prediction, their annotation process is labor-intensive and time-consuming, resulting in substantial costs. Consequently, trimap-based methods face challenges in widespread adoption in industrial applications.
%

To overcome these limitations, researchers~\cite{yang2022exploring, yu2021mask, wei2021improved, ye2024unifying, xia2024towards} have proposed interactive matting, which replaces trimaps with simpler and more accessible auxiliary inputs, such as points, bounding boxes, or masks.
The success of large pre-trained segmentation models, such as SAM~\cite{kirillov2023segment, ravi2024sam, ke2023segment}, has propelled the advancement of numerous downstream tasks, including interactive matting. A series of SAM-based matting methods~\cite{li2024matting, yao2024matte, xia2024towards} utilizes stacked modules to progressively refine SAM-generated masks, thereby producing more precise alpha mattes. However, these methods often freeze SAM during training, which prevents them from correcting errors in SAM's output. As a result, any inaccuracies in SAM’s output are amplified by subsequent stacked modules, leading to inaccurate alpha matte predictions.

Recently, diffusion models~\cite{ho2020denoising, song2020denoising, rombach2022high, podell2023sdxl, esser2024scaling} have achieved significant success in the field of image generation, demonstrating great application and research value.
By training on billions of text-image pairs, diffusion models achieve robust generalization, providing universal image representations while maintaining fine-detail preservation. These outstanding characteristics make it a promising candidate for various visual perception tasks.
For example, Marigold~\cite{ke2024repurposing} demonstrates that diffusion models, even when fine-tuned only on synthetic datasets, can achieve remarkable performance in depth estimation, thanks to their strong generalization and detail-preserving capabilities.
%
%
Building on this, extensive studies~\cite{tian2024diffuse, karmann2024repurposing, amit2021segdiff, ji2023ddp, zavadski2024primedepth, zhang2024betterdepth, ye2024diffusionedge, hu2024diffumatting, wang2024matting, zhang2025ar} have further explored the potential of diffusion models in image perception tasks, making them an effective paradigm for various downstream tasks, including interactive image matting.

Although diffusion models demonstrate strong potential in visual perception tasks, most existing approaches fine-tune them with empty text embeddings, which compromises their robust text-driven interaction capabilities.
To address this issue, we present SDMatte, a diffusion-based interactive matting method that leverages the powerful priors of diffusion models while fully exploiting their interactive capabilities.
Specifically, we follow a one-step deterministic paradigm similar to GenPercept~\cite{xu2024matters}, and enhance it by introducing visual prompts (points, boxes, and masks) to enable interactive matting.
First, we propose a visual prompt-driven cross-attention mechanism, which effectively inherits the powerful text-driven interaction capability of diffusion models and transforms it into a visual prompt-driven interaction capability.
Additionally, we integrate the coordinate embeddings of visual prompts and the opacity embeddings of target objects into the U-Net of the diffusion model, enhancing the model's sensitivity to spatial position and opacity information. 
Finally, we design a masked self-attention mechanism, which allows the model to focus more on the regions specified by the visual prompts, thereby improving performance. Our contributions can be summarized as follows:
\begin{itemize}
    \item We propose SDMatte, which harnesses the powerful priors of diffusion models and transforms their text-driven interaction capability into visual prompt-driven interaction capability through a visual prompt-driven cross-attention mechanism, facilitating interactive matting.
    \vspace{4pt}
    
    \item We significantly enhance the model’s sensitivity to spatial position and opacity information by integrating coordinate embeddings and opacity embeddings into the U-Net architecture of the diffusion model.
    \vspace{4pt}

    \item We propose a masked self-attention mechanism, enabling the model to focus more on the regions specified by the visual prompts, thereby enhancing performance.
    \vspace{4pt}

    \item Extensive evaluations on various benchmarks, including AIM-500~\cite{li2021deep}, AM-2k~\cite{li2022bridging}, P3M~\cite{li2021privacy} and RefMatte~\cite{li2023referring}, demonstrate that SDMatte can achieve superior performance compared to existing interactive matting methods, while also exhibiting robust generalization capabilities.
\end{itemize}

\section{Related Work}
\label{sec:related work}
\subsection{Interactive Matting}
Image matting~\cite{chen2022pp, yu2021mask, ke2022modnet, wang2024matting, park2023mask, hu2024diffusion, huynh2024maggie, sun2024semantic, hu2024diffumatting, guo2024context} has attracted extensive research interest in recent years, which can be mainly divided into three categories, including trimap-based approaches~\cite{forte2020f, yao2024vitmatte, hu2025diffusion, jiang2023trimap, zhou2023sampling}, automatic matting approaches~\cite{li2021deep, li2022bridging, li2021privacy, ma2023rethinking, ye2024unifying}, and interactive matting approaches~\cite{yu2021mask, wei2021improved, li2024matting, yao2024matte, ye2024unifying, xia2024towards}.
The trimap-based approaches can achieve high-quality matting results but often require substantial human effort to obtain trimaps. The automatic matting approaches aim to predict the alpha matte without any auxiliary inputs but often produce unsatisfactory results for non-salient and transparent objects.
Our method falls into the interactive matting category, which aims to extract accurate alpha mattes based on simple visual prompts (e.g., points, boxes, and masks) provided by users. 

Recently, the emergence of SAM~\cite{kirillov2023segment, ravi2024sam, ke2023segment} has advanced a variety of downstream tasks, including interactive matting. MAM~\cite{li2024matting} refines the coarse masks produced by SAM into fine-grained alpha mattes by appending a lightweight mask-to-matte module to the frozen SAM. MatAny~\cite{yao2024matte} integrates existing models, including SAM~\cite{kirillov2023segment}, to extract alpha mattes in a training-free manner. SEMat~\cite{xia2024towards} proposes a matte-aligned decoder and novel training objectives to convert the coarse masks into high-quality alpha mattes. 
%
However, these methods typically depend heavily on SAM. As a result, errors in SAM’s output are propagated and amplified by the subsequent modules, leading to inaccurate alpha matte predictions.
In contrast, SmartMatting~\cite{ye2024unifying} abandons the heavy interactive mechanism of SAM in favor of a more lightweight interaction design, but struggles to handle objects with rich fine-grained details.

\subsection{Diffusion Models for Visual Perception}
Diffusion models~\cite{ho2020denoising, song2020denoising, mukhopadhyay2023diffusion, ho2022classifier, saharia2022palette, rombach2022high, podell2023sdxl, esser2024scaling} have recently achieved remarkable success in image generation. They generate high-fidelity and fine-grained images through a unique process of noise addition and denoising.
The remarkable achievements of diffusion models in image generation have motivated researchers to explore their potential in visual perception tasks such as segmentation, depth estimation, etc. This motivation stems from the fact that diffusion models are trained on large-scale datasets, enabling them to provide strong prior knowledge.
Marigold~\cite{ke2024repurposing} first leverages the strong priors of diffusion models for monocular depth estimation, which surpasses CNN-based and Transformer-based approaches in both accuracy and generalization, even with fine-tuning solely on synthetic datasets.
DAS~\cite{tian2024diffuse} and M2N2~\cite{karmann2024repurposing} propose unsupervised zero-shot segmentation frameworks by exploiting the intrinsic priors of attention layers in diffusion models.
DiffDIS~\cite{yu2024high} leverages the pre-trained U-Net of diffusion models to directly generate high-resolution, fine-grained segmentation masks in a single step.
GenPercept~\cite{xu2024matters} proposes a one-step deterministic paradigm that eliminates the denoising process. Instead, it directly supervises prediction maps in the pixel space, thereby accelerating inference and reducing erroneous detail generation.
Furthermore, DiffuMatting~\cite{hu2024diffumatting} fully exploits diffusion models combined with a green screen design to achieve efficient data annotation and controllable generation.
MbG~\cite{wang2024matting} reformulates image matting as a generative modeling problem using diffusion models, enabling fine-grained alpha matte prediction.

Although these works fully exploit the strong priors of diffusion models and achieve substantial progress, they often overlook or even undermine the powerful interactive capabilities of diffusion models. 
In this paper, we present SDMatte for interactive matting.
SDMatte leverages the powerful priors of diffusion models and transforms the text-driven interaction capabilities into more suitable visual prompt-driven interaction capabilities for interactive matting, fully exploiting the potential of diffusion models.
\begin{figure*}[!htbp]
    \centering
    \includegraphics[width=0.9\linewidth]{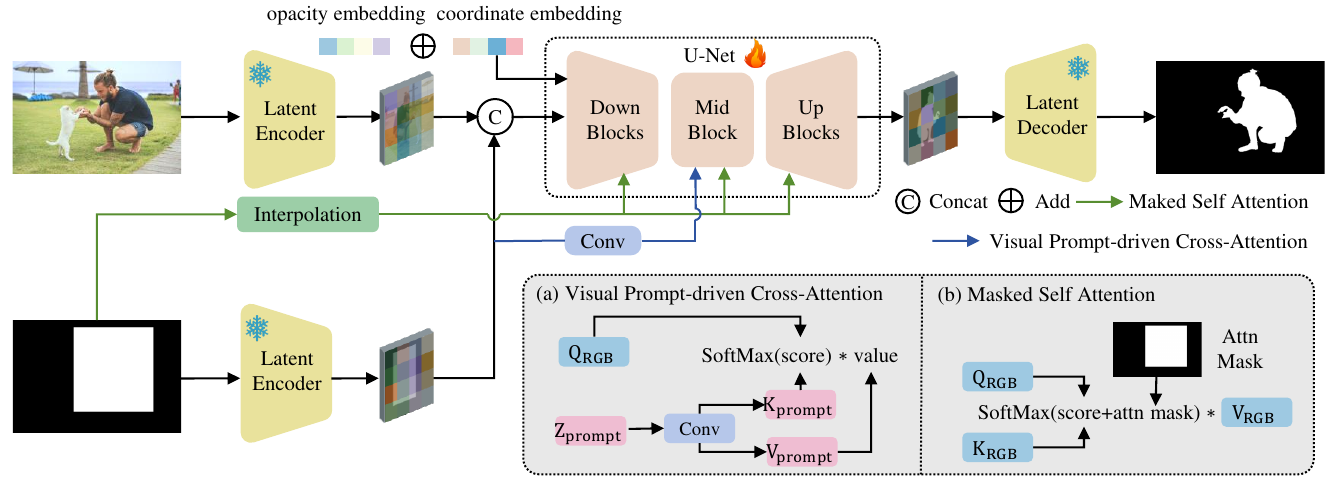}
    \vspace{-5pt}
    \caption{\textbf{The overall framework of SDMatte.} 
    We map the input image and visual prompt into the latent space and concatenate them as the input to the U-Net. Subsequently, we substitute the time embedding in Stable Diffusion with coordinate embeddings of visual prompts and opacity embeddings of target objects to enhance SDMatte’s sensitivity to spatial position and opacity information. Finally, we leverage the masked self-attention and visual prompt-driven cross-attention mechanisms to maximize the effectiveness of visual prompts, guiding the U-Net in generating the alpha matte and map it back to pixel space.}
    \label{fig:pipe}
    \vspace{-5pt}
\end{figure*}

\section{Methodology}
\subsection{Overall Paradigm}
\label{sec:overall}
To address the limitations of existing interactive matting methods in capturing intricate edge details, we propose SDMatte, a diffusion-driven interactive matting model that fully exploits the exceptional properties of diffusion models, including strong prior knowledge, superior detail preservation capabilities, and robust text-driven interaction capabilities.

As shown in \figref{fig:pipe}, our approach is based on Stable Diffusion v2~\cite{rombach2022high} for interactive image matting.
Specifically, we first employ the VAE encoder to map the input image and visual prompts from the pixel space into the latent space. 
Subsequently, the latent representations of the input image and visual prompts are concatenated and passed into the U-Net. To accommodate the increased input dimensions, the first-layer convolutional weights of the U-Net are duplicated.
Finally, we utilize the VAE decoder to remap the U-Net’s output to the pixel space for matting loss computation and supervision.
As image matting aims to predict boundary transparency, the stochasticity property of diffusion models hinders their performance in predicting alpha map.
Thus, we adopt the one-step deterministic paradigm and remove the noise addition and denoising process.

However, diffusion models are inherently powerful text-driven frameworks for interactive image generation, while merely concatenating image and visual prompts in the latent space fails to fully exploit their interactive potential.
To inherit the powerful text-driven interaction capability of diffusion models and transform it into visual prompt-driven interaction capability, we propose a visual prompt-driven cross-attention mechanism, which will be elaborated in \secref{sec:cross_attn}. 
To enhance SDMatte’s sensitivity to spatial position information and object opacity information, we introduce coordinate embedding and opacity embedding, which will be elaborated in \secref{sec:emb}. 
To improve the model’s attention to regions indicated by visual prompts, we propose a masked self-attention mechanism depicted in \secref{sec:masked}.

\subsection{Visual Prompt Cross-Attention Mechanism}
\label{sec:cross_attn}
Although diffusion models possess powerful text-driven interaction capability, abstract text embedding struggles to provide accurate location information guiding the extraction of alpha matte. 
Therefore, we propose a visual prompt-driven cross-attention mechanism, which inherits the text-driven interactive capability of diffusion models and translates it into a visual prompt-driven interactive capability.
%
%
This mechanism replaces the original text embedding with a visual prompt embedding and projects it to the same dimension as the text embedding to facilitate weight reuse in the cross-attention layer.

\begin{figure}[tbp]
    \centering
    \includegraphics[width=0.99\linewidth]{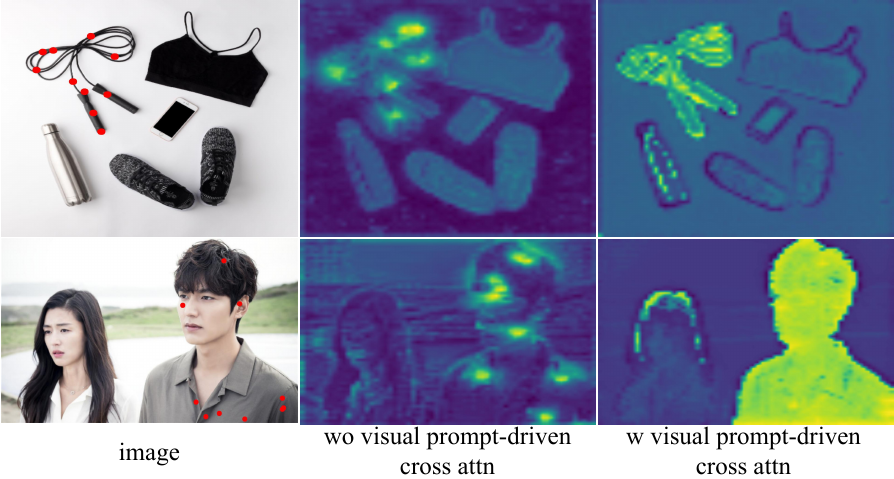}
    \vspace{-5pt}
    \caption{\textbf{Visualization of the attention maps in U-Net's final cross-attention layer.} It visually demonstrates the model's focus on the regions indicated by the visual prompts, proving the effectiveness of the visual prompt-driven cross-attention mechanism. }  
    \label{fig:attn_map}  
    \vspace{-5pt}
\end{figure}

Specifically, as shown in \figref{fig:pipe}a, we apply a zero convolution layer to map the latent representation of the visual prompt to the same dimension as the text embedding. It is subsequently used to replace the text embedding in the diffusion model and is fed into the cross-attention module of the U-Net's middle block, where semantic information is most concentrated.
The pre-trained weight of the text-driven interaction mechanism and the unique design of zero convolution layer enable the visual prompt-driven cross-attention mechanism to gradually convert the text-driven interaction capability of diffusion model into visual prompt-driven interaction capability during training.
As depicted in \figref{fig:attn_map}, the visual prompt embedding provides SDMatte with more precise location information compared to text embedding. This strongly validates the effectiveness of the visual prompt-driven cross-attention mechanism.

\subsection{Opacity and Coordinate Embeddings}
\label{sec:emb}
In SDXL~\cite{podell2023sdxl}, image size and cropping coordinates are used as conditions of the U-Net, which are encoded as embeddings and added to the time embedding.
This design drives the model to learn the image resolution and cropping position information, which allows the model to adapt to various image sizes during the inference phase while ensuring that the generated patterns remain centered.
Inspired by this, we introduce the coordinate information and opacity information of target objects as a condition to guide the generation of alpha matte, enhancing model's sensitivity to spatial position and opacity of target objects.
Additionally, in diffusion models, the time embedding represents the level of noise added at each timestep. However, it is useless in our deterministic paradigm, so we empirically remove it.

Specifically, for the box prompt, we apply sinusoidal positional encoding to the coordinates of the top-left and bottom-right corners. Each of the four numbers is encoded into a $C/4$-dimensional vector, resulting in $\mathbf{E}_{box} \in \mathbb{R}^{B \times C}$. For the mask prompt, we first compute the minimal bounding box that can enclose the mask, and then encode it using the same strategy as the box prompt. For $N$ point prompts, we first check whether $2N$ is divisible by $C$. If not, we add $P$ zeros to the coordinate list such that $2N + P$ becomes divisible by $C$. Subsequently, we apply sinusoidal positional encoding to the $2N+P$ numbers, resulting in $\mathbf{E}_{point}\in \mathbb{R}^{B \times C}$.

\begin{equation}
    C = \begin{cases}
    1680,\hspace{0.2cm} & \text{point prompt}\\
    1280, & \text{box or mask prompt}
    \end{cases}
    \label{eq:coor_emb}
\end{equation}

\noindent Here, the values of $C_{box}$ and $C_{mask}$ are determined according to the time embedding configuration in diffusion models, in which a scalar is mapped to a 320-dimensional vector. For $C_{point}$, it is chosen such that it can be divisible by most prime numbers, thereby minimizing $P$.

In the field of image matting, the extraction of alpha mattes for transparent objects remains a significant challenge. To enhance SDMatte's ability to recognize transparent objects, we annotate all training and testing data with opacity information. If an object is transparent, its opacity is set to 0; otherwise, it is set to 1. Subsequently, we also apply sinusoidal positional encoding to the object's opacity information to produce $\mathbf{E}_{opacity}$. Finally, we use a linear combination of opacity embedding and coordinate embedding as a substitute for the time embedding in diffusion models:

\begin{equation}
    \mathbf{E}_{cond} = f_1(\mathbf{E}_{opacity}) + f_2(\mathbf{E}_{coord}).
    \label{eq:emb}
\end{equation}
Here, $f_1$ and  $f_2$ represent linear layers.

\subsection{Masked Self-Attention Mechanism}
\label{sec:masked}
Although the self-attention mechanism in diffusion models performs global dependency modeling, it fails to explicitly prioritize prompt-indicated regions, 
which constrains the model’s potential to leverage visual prompts effectively.
In Mask2Former~\cite{cheng2022masked}, the masked cross-attention mechanism is designed to focus only on the foreground region of each query’s predicted mask, thereby accelerating the convergence of Transformer-based models.
Inspired by this, we propose a masked self-attention mechanism that enables the model to focus more effectively on the regions indicated by visual prompts while disregarding irrelevant areas, thereby fully leveraging the potential of visual prompts.

Specifically, for box and mask prompts, we generate hard binary attention masks $\mathbf{M}_b \in \{ 0,1\}$ and $\mathbf{M}_m \in \{ 0,1\}$, which explicitly indicate the regions where the model should allocate more attention, as defined by:

\begin{equation}
    \mathbf{M}_{(x,y)} = \begin{cases}
    1, & \text{if  }(x,y) \in \text{region}  \\
    0. & \text{otherwise}
    \end{cases}
    \label{eq:masked}
\end{equation}
For point prompts, we generate a soft attention mask $\mathbf{M}_p \in[0,1]$ centered at the point coordinates, which follows a standard normal distribution to smoothly weight the surrounding regions.
As shown in \figref{fig:pipe}b, the attention mask modulates the attention map as follows:
\begin{equation}
    \begin{split}
        \mathbf{M} &= (\mathbf{M}-1)*\infty \\
        \mathbf{X} &= \text{softmax}(\mathbf{M}+\frac{\mathbf{Q}\mathbf{K}^T}{\sqrt{d_k}})\mathbf{V}.
    \end{split}
    \label{eq:score}
\end{equation}
\noindent Here, $\mathbf{Q}$ denotes query, $\mathbf{K}$ denotes key, $\mathbf{V}$ denotes value and $\mathbf{X}$ denotes the input to the subsequent layer. This mechanism dynamically adjusts the model’s attention according to visual prompts, leading to improved performance in interactive scenarios driven by prompts.


\section{Experiments}
\begin{table*}[htbp]
    \centering
    \renewcommand{\arraystretch}{1}
    \setlength{\tabcolsep}{2mm}{
    \resizebox{\textwidth}{!}{
    \begin{tabular}{c|c|c|cccccc|cccccc}
        \toprule
        \multirow{2}{*}{Method} & \multirow{2}{*}{\shortstack{Pretrained \\ Backbone}} & \multirow{2}{*}{Prompt} & \multicolumn{6}{c|}{AIM-500 (natural)} & \multicolumn{6}{c}{AM-2K (animal)}  \\ 
         &  &  & MSE $\downarrow$ & MAD $\downarrow$ & SAD $\downarrow$ & Grad $\downarrow$ & Conn $\downarrow$ & Impro $\uparrow$ & MSE $\downarrow$ & MAD $\downarrow$ & SAD $\downarrow$ & Grad $\downarrow$ & Conn $\downarrow$ & Impro $\uparrow$ \\ 
        \midrule
        MAM~\cite{li2024matting} & SAM & point & 0.0752 & 0.1080 & 186.50 & 37.48 & 40.38 & -120.86\% & 0.0597 & 0.0813 & 141.60 & 22.48 & 31.52 & -82.06\% \\
        
        MatAny~\cite{yao2024matte} & SAM & point & 0.0425 & 0.0523 & 87.05 & 33.44 & 25.35 & -22.73\% & 0.0116 & 0.0188 & 32.20 & 15.68 & 20.39 & 36.89\% \\
        
        SmartMatting~\cite{ye2024unifying} & DINOv2 & point & 0.0302 & 0.0388 & 66.27 & 46.63 & \colorbox{second}{18.77} & - & 0.0302 & 0.0366 & 62.61 & 33.82 & \colorbox{second}{15.93} & -\\

        LiteSDMatte & SD2 & point & \colorbox{second}{0.0115} & \colorbox{second}{0.0207}  & \colorbox{second}{34.43} & \colorbox{first}{24.32} & 19.97 & \colorbox{second}{39.61\%} & \colorbox{second}{0.0095} & \colorbox{second}{0.0161} & \colorbox{second}{27.51} & \colorbox{second}{13.59} & 17.74 & \colorbox{second}{45.81\%}\\
        
        SDMatte & SD2 & point & \colorbox{first}{0.0109} & \colorbox{first}{0.0189} & \colorbox{first}{31.80} & \colorbox{second}{26.84} & \colorbox{first}{17.51} & \colorbox{first}{43.27\%} & \colorbox{first}{0.0060} & \colorbox{first}{0.0104} & \colorbox{first}{17.54} & \colorbox{first}{13.17} & \colorbox{first}{10.86} & \colorbox{first}{63.32\%}\\
        
        \midrule
        MAM~\cite{li2024matting} & SAM & box & 0.0116 & 0.0222 & 36.66 & {21.04} & 18.99 & {-32.02\%} & 0.0038 & 0.0100 & 17.14 & {11.28} & 10.34 & {-1.58\%}\\
        
        MatAny~\cite{yao2024matte} & SAM & box & 0.0545 & 0.0640 & 106.26 & 31.74 & 20.24 & -263.50\% & 0.0136 & 0.0204 & 35.30 & 14.07 & 17.57 & -120.06\%\\
        
        SmartMatting~\cite{ye2024unifying} & DINOv2 & box & {0.0077} & {0.0151} & {25.33} & 27.16 & {13.54} & - & {0.0038} & {0.0088} & {14.91} & 16.53 & {9.31} & -\\

        SEMat~\cite{xia2024towards} & SAM2 & box & {0.0071} & {0.0146} & {24.30} & \colorbox{second}{16.06} & {13.64} & 11.06\% & \colorbox{second}{0.0028} & {0.0075} & {12.89} & \colorbox{second}{8.69} & {8.44} & 22.28\% \\

        LiteSDMatte & SD2 & box & 0.0056 & 0.0124  & 20.83 & 20.94 & 12.90 & 18.11\% & 0.0033 & 0.0073 & 12.54 & 11.08 & 8.49 & 17.58\% \\
        
        SDMatte & SD2 & box & \colorbox{second}{0.0049} & \colorbox{second}{0.0116} & \colorbox{second}{19.45} & {20.63} & \colorbox{second}{12.58} & \colorbox{second}{22.78\%} & {0.0029} & \colorbox{second}{0.0065} & \colorbox{second}{11.04} & {10.09} & \colorbox{second}{6.99} & \colorbox{second}{27.93\%}\\

        $\text{SDMatte}^{*}$ & SD2 & box & \colorbox{first}{0.0036} & \colorbox{first}{0.0097} & \colorbox{first}{16.42} & \colorbox{first}{14.89} & \colorbox{first}{11.00} & \colorbox{first}{37.62\%} & \colorbox{first}{0.0020} & \colorbox{first}{0.0054} & \colorbox{first}{9.23} & \colorbox{first}{8.69} & \colorbox{first}{6.41} & \colorbox{first}{40.54\%}\\
        \midrule
        MGMatting~\cite{yu2021mask} & - & mask & 0.0155 & 0.0285 & 48.28 & 20.78 & 20.26 & - & 0.0199 & 0.0309 & 53.31 & 10.92 & 13.95 & - \\

        LiteSDMatte & SD2 & mask & \colorbox{second}{0.0030} & \colorbox{second}{0.0094}  & \colorbox{second}{15.83} & \colorbox{second}{19.17} & \colorbox{second}{11.29} & \colorbox{second}{53.38\%} & \colorbox{second}{0.0014} & \colorbox{second}{0.0049} & \colorbox{second}{8.45} & \colorbox{second}{9.55} & \colorbox{second}{6.57} & \colorbox{second}{65.34\%}\\

        SDMatte & SD2 & mask & \colorbox{first}{0.0027} & \colorbox{first}{0.0087} & \colorbox{first}{14.53} & \colorbox{first}{16.94} & \colorbox{first}{10.95} & \colorbox{first}{57.28\%} & \colorbox{first}{0.0012} & \colorbox{first}{0.0043} & \colorbox{first}{7.30} & \colorbox{first}{6.96} & \colorbox{first}{5.78} & \colorbox{first}{72.24\%}\\
        
        \midrule
        \multicolumn{3}{@{}c@{}}{} & \multicolumn{6}{|c|}{P3M-500-NP (human)} & \multicolumn{6}{c}{RefMatte-RW-100 (human)} \\
        \midrule
        MAM~\cite{li2024matting} & SAM & point & 0.0875 & 0.1163 & 207.53 & 29.43 & 43.49 & -200.35\% & 0.1651 & 0.1896 & 336.49 & 49.91 & 27.80 & -806.15\% \\

        MatAny~\cite{yao2024matte} & SAM & point & 0.0295 & 0.0342 & 57.33 & 25.95 & \colorbox{first}{15.97} & -5.37\% & 0.0118 & 0.0137 & 24.35 & 18.13 & \colorbox{second}{4.98} & \colorbox{second}{11.03\%}\\
        
        SmartMatting~\cite{ye2024unifying} & DINOv2 & point & 0.0239 & 0.0291 & 50.46 & 28.50 & \colorbox{second}{19.64} & - & 0.0127 & 0.0153 & 26.75 & 23.01 & 5.38 & -\\

        LiteSDMatte & SD2 & point & \colorbox{first}{0.0121} & \colorbox{first}{0.0173}  & \colorbox{first}{29.94} & \colorbox{first}{16.55} & 21.82 & \colorbox{first}{32.28\%} & \colorbox{second}{0.0096} & \colorbox{second}{0.0131} & \colorbox{second}{22.90} & \colorbox{second}{15.74} & 7.29 & 9.85\% \\
        
        SDMatte & SD2 & point & \colorbox{second}{0.0134} & \colorbox{second}{0.0183} & \colorbox{second}{32.02} & \colorbox{second}{20.35} & 20.76 & \colorbox{second}{28.10\%} & \colorbox{first}{0.0091} & \colorbox{first}{0.0116} & \colorbox{first}{20.45} & \colorbox{first}{15.57} & \colorbox{first}{4.01} & \colorbox{first}{26.78\%} \\

        \midrule
        MAM~\cite{li2024matting} & SAM & box & 0.0061 & 0.0115 & 18.86 & {13.58} & {9.56} & {-21.81\%} & 0.0124 & 0.0179 & 31.46 & 15.93 & 5.45 & 14.03\%\\
        
        MatAny~\cite{yao2024matte} & SAM & box & 0.0328 & 0.0372 & 60.97 & 22.22 & 13.62 & -306.77\% & {0.0118} & {0.0136} & {23.85} & {15.63} & {4.47} & {27.66\%}\\
        
        SmartMatting~\cite{ye2024unifying} & DINOv2 & box & {0.0037} & {0.0081} & {14.10} & 18.31 & 10.14 & - & 0.0173 & 0.0199 & 34.86 & 23.86 & 4.90 & - \\

        SEMat~\cite{xia2024towards} & SAM2 & box & {0.0028} & {0.0063} & {10.88} & 11.19 & {7.67} & 26.53\% & {0.0055} & {0.0075} & {13.24} & \colorbox{second}{10.58} & {3.12} & 56.90\%\\

        LiteSDMatte & SD2 & box & 0.0025 & 0.0054  & 9.31 & 12.56 & 6.83 & 32.76\% & 0.0060 & 0.0082 & 14.39 & 12.85 & 3.58 & 51.18\%\\
        
        SDMatte & SD2 & box & \colorbox{second}{0.0020} & \colorbox{second}{0.0046} & \colorbox{second}{7.90} & \colorbox{first}{9.32} & \colorbox{second}{6.31} & \colorbox{second}{44.00\%} & \colorbox{second}{0.0047} & \colorbox{second}{0.0062} & \colorbox{second}{10.92} & {11.41} & \colorbox{first}{2.80} & \colorbox{second}{61.08\%} \\

        $\text{SDMatte}^{*}$ & SD2 & box & \colorbox{first}{0.0016} & \colorbox{first}{0.0044} & \colorbox{first}{7.58} & \colorbox{second}{10.87} & \colorbox{first}{5.85} & \colorbox{first}{46.32\%} & \colorbox{first}{0.0041} & \colorbox{first}{0.0059} & \colorbox{first}{10.33} & \colorbox{first}{10.54} & \colorbox{first}{2.41} & \colorbox{first}{64.73\%}\\
        \midrule
        MGMatting~\cite{yu2021mask} & - & mask & 0.0100 & 0.0178 & 30.48 & 14.93 & 13.40 & - & 0.0258 & 0.0326 & 56.06 & 16.17 & 9.56 & - \\

        LiteSDMatte & SD2 & mask & \colorbox{second}{0.0011} & \colorbox{second}{0.0039}  & \colorbox{second}{6.66} & \colorbox{second}{11.10} & \colorbox{second}{5.22} & \colorbox{second}{66.39\%} & \colorbox{second}{0.0009} & \colorbox{second}{0.0022} & \colorbox{second}{3.86} & \colorbox{second}{8.44} & \colorbox{second}{2.31} & \colorbox{second}{81.30\%}\\

        SDMatte & SD2 & mask & \colorbox{first}{0.0007} & \colorbox{first}{0.0030} & \colorbox{first}{5.10} & \colorbox{first}{6.47} & \colorbox{first}{4.12} & \colorbox{first}{77.07\%} & \colorbox{first}{0.0008} & \colorbox{first}{0.0019} & \colorbox{first}{3.27} & \colorbox{first}{6.23} & \colorbox{first}{1.88} & \colorbox{first}{85.41\%}\\
        \bottomrule
    \end{tabular}}}
    \vspace{-0pt}
    \caption{\textbf{Performance comparison with existing interactive image matting methods.} The results are produced using the official models provided by the authors without any retraining. The \colorbox{first}{text} represents the best method, and the  \colorbox{second}{text} represents the second-best method. “Impro” denotes the average relative improvement on the five metrics compared with the baseline SmartMatting. $\text{SDMatte}^{*}$ is a version trained on set 2, using box prompt for guidance. It is used for comparison with SEMat, which only supports box prompt. }
    \label{table:main_result}
    \vspace{-5pt}
\end{table*}

 \begin{figure*}[t]
    \centering
    \includegraphics[height=11.5cm, width=0.98\linewidth]{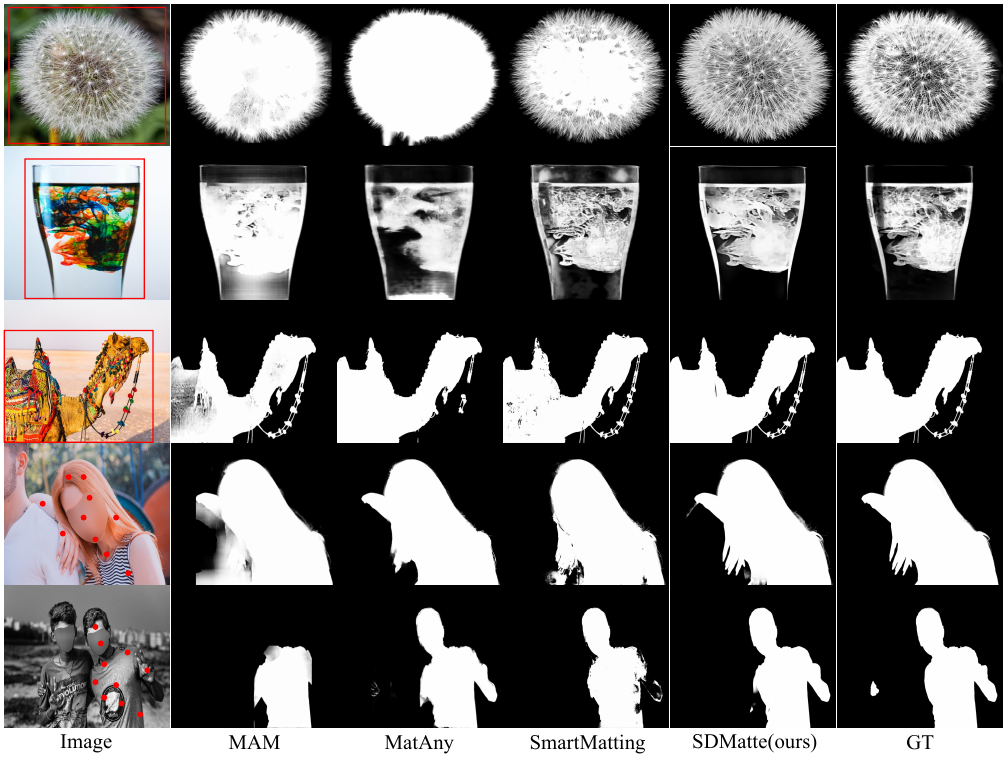}
    \vspace{-8pt}
    \caption{\textbf{Visual comparison with existing interactive image matting methods.} Compared to other methods, our approach demonstrates significantly better generalization and superior extraction capabilities for transparent and detail-rich objects.}  
    \label{fig:compare}  
    \vspace{-5pt}
\end{figure*}

\subsection{Implementation Details}
\noindent \textbf{Datasets:} We adopt the same training set as SmartMatting~\cite{ye2024unifying}, which includes Composition-1k~\cite{xu2017deep}, Distinctions-646~\cite{qiao2020attention}, AM-2k~\cite{li2022bridging}, UHRSD~\cite{xie2022pyramid}, and 10000 images from RefMatte~\cite{li2023referring}, denoted as set 1.
Additionally, recent work SEMat~\cite{xia2024towards} proposes a large-scale dataset of real human portraits, named COCO-Matte. To enable a comprehensive comparison, we also adopt the same training set as SEMat, which includes Composition-1k~\cite{xu2017deep}, Distinctions-646~\cite{qiao2020attention}, AM-2k~\cite{li2022bridging}, and COCO-Matte~\cite{xia2024towards}, denoted as set 2.

\noindent \textbf{Benchmarks and Metrics:} We evaluate our method across a diverse set of image matting benchmarks, including AIM-500~\cite{li2021deep}, AM-2k ~\cite{li2022bridging}, P3M~\cite{li2021privacy} and RefMatte-RW-100~\cite{li2023referring}. To measure the quality of the predicted alpha matte, we employ five standard metrics: MSE, MAD, SAD, Grad~\cite{rhemann2009perceptually} and Conn~\cite{rhemann2009perceptually}.

\noindent \textbf{Training Details:} The SDMatte model is optimized using the AdamW optimizer with a learning rate of \(1 \times e^{-4}\). The model is trained for 50 epochs on two NVIDIA H20 GPUs, with a batch size of 9 per GPU. For the learning rate scheduler, we employ a warmup strategy combined with an exponential decay scheduler. We initialize SDMatte with the pre-trained weights of Stable Diffusion v2 and adopt a mixed prompt strategy during training, where point, bounding box, and mask prompts are randomly generated for each sample. We perform a foreground duplication strategy with a 50\% probability. Specifically, for each synthesized image, the foreground object without any prompt is duplicated alongside the prompted one on the same background, thereby enhancing the model's sensitivity to visual prompts.

\begin{table}[htbp]
    \centering
    \renewcommand{\arraystretch}{1}
    \setlength{\tabcolsep}{4mm}{
    \resizebox{0.88\linewidth}{!}{
    \begin{tabular}{c|c|c|c}
        \toprule
        Method & Parameters (M) & FLOPs (G) & Latency (ms) \\ 
         \midrule
          MAM & 644 & 3055 & 454 \\
          MatAny & 910 & 3948 & 655 \\
          Smat & 27 & 538 & 190 \\ \hline
          SDMatte & 957 & 11203 & 1014 \\
          LiteSDMatte & 593 & 2010 & 366 \\
        \bottomrule
    \end{tabular}}}
    \vspace{-0pt}
        \caption{Comprehensive comparison of computational complexity with existing methods. All reported results are derived from inference conducted on 1K resolution images on H20.}
    \label{table:complexity}
    \vspace{-5pt}
\end{table}

\begin{table*}[htbp]
    \centering
    \renewcommand{\arraystretch}{1}
    \setlength{\tabcolsep}{2mm}{
    \resizebox{0.9\textwidth}{!}{
    \begin{tabular}{c|c|c|cc|cc|cc|cc|c}
        \toprule
        \multirow{2}{*}{\shortstack{Down\\Blocks}} & \multirow{2}{*}{\shortstack{Mid\\Block}} & \multirow{2}{*}{\shortstack{Up\\Blocks}} & \multicolumn{2}{c|}{AIM-500 (point)} & \multicolumn{2}{c|}{RefMatte-RW-100(point)} & \multicolumn{2}{c|}{AIM-500 (box)} & \multicolumn{2}{c|}{RefMatte-RW-100(box)} & \multirow{2}{*}{\shortstack{Impro $\uparrow$}} \\ 
         &  &  & MSE $\downarrow$ & SAD $\downarrow$ & MSE $\downarrow$ & SAD$\downarrow$ & MSE $\downarrow$ & SAD $\downarrow$ & MSE $\downarrow$ & SAD $\downarrow$  \\ 
         \midrule
          &  &  & 0.0135 & 40.53 & 0.0156 & 36.56 & 0.0087 & 25.79 & 0.0061 & 15.74 & - \\
          \checkmark &  &  & 0.0122 & 39.50 & 0.0162 & 36.88 & 0.0111 & 29.76 & 0.0060 & 15.52 & -4.06\% \\
          & \checkmark &  & 0.0111 & 38.02 & 0.0135 & 34.07 & 0.0070 & 24.01 & 0.0053 & 14.23 & \colorbox{first}{11.67\%} \\
          &  & \checkmark & 0.0140 & 42.57 & 0.0149 & 38.27 & 0.0103 & 28.75 & 0.0068 & 17.61 & -7.77\% \\
          \checkmark & \checkmark &  & 0.0127 & 40.77 & 0.0146 & 36.12 & 0.0062 & 21.94 & 0.0066 & 16.72 & 5.27\% \\
          \checkmark &  & \checkmark & 0.0174 & 49.07 & 0.0166 & 37.38 & 0.0094 & 27.65 & 0.0078 & 19.16 & -15.43\%\\
          & \checkmark & \checkmark & 0.0147 & 44.70 & 0.0154 & 38.03 & 0.0087 & 25.92 & 0.0084 & 20.44 & -11.25\% \\
         \checkmark & \checkmark & \checkmark & 0.0154 & 44.31 & 0.0184 & 41.89 & 0.0061 & 20.23 & 0.0062 & 15.60 & -0.65\% \\
        \bottomrule
    \end{tabular}}}
    \vspace{-5pt}
    \caption{\textbf{Ablation of Visual Prompt-driven Cross-Attention Mechanism.} We apply the visual prompt-driven cross-attention mechanism in various modules of the SDMatte to evaluate its sensitivity across different modules and identify the optimal performance setting. The baseline is set as the configuration without visual prompt-driven cross-attention mechanism. }
    \label{table:cross attn}
    \vspace{-5pt}
\end{table*}

\subsection{Main Results}
In this section, we compare our method with previous state-of-the-art approaches, such as MatAny~\cite{yao2024matte}, MAM~\cite{li2024matting}, SmartMatting~\cite{ye2024unifying} and SEMat~\cite{xia2024towards} from two
aspects: performance and efficiency, to validate the effectiveness of SDMatte in the interactive image matting task.

\noindent \textbf{Overall Performance Comparison: }
As shown in \tabref{table:main_result}, we perform a comprehensive comparison of our method with existing state-of-the-art methods based on other pre-trained weights, including SAM~\cite{kirillov2023segment} and DINOv2~\cite{oquab2023dinov2}. 
Notably, for SDMatte's mask prompt mode, since the classic work MGMat-wild~\cite{park2023mask} has not been publicly released, we compare it with the older work MGMatting~\cite{yu2021mask}.
On the AIM-500 benchmark, which contains foreground objects from diverse categories, our method surpasses all comparison methods, demonstrating superior generalization across diverse categories.
On the AM-2K benchmark, which only contains animal foregrounds, and the P3M-500-NP benchmark, which emphasizes portrait foregrounds, our method outperforms all comparative methods, demonstrating superior performance on common foreground objects.
On the multi-person benchmark RefMatte-RW-100, our method also exceeds all comparative methods, demonstrating greater sensitivity to visual prompts.
Furthermore, as shown in \figref{fig:compare}, we provide a visual comparison with other interactive image matting methods. Compared to previous methods, SDMatte fully leverages the powerful priors of the Stable Diffusion model, achieving better detail generation. Our method exhibits remarkable robustness across various types of visual prompts, consistently yielding accurate alpha matte predictions.

\noindent \textbf{Efficiency Comparison with Other Methods: }
Although our method can achieve excellent results, we notice that diffusion-based models will bring more heavier computational burden than other matting methods, which may limit the applicability of SDMatte in practice.
To address this limitation, we implement a lightweight variant named LiteSDMatte. Specifically, we construct LiteSDMatte by replacing the VAE and U-Net in SDMatte with TinyVAE~\cite{taesd} and the base version of BK-U-Net~\cite{kim2024bk} to achieve a more lightweight architecture.
As shown in \tabref{table:complexity}, LiteSDMatte achieves a significant improvement in computational efficiency, outperforming all SAM-based methods and being only slower than the lightweight SmartMatting approach.
Additionally, we perform feature-level aligned distillation on LiteSDMatte, enabling it to inherit the strong interactive matting capability of SDMatte while preserving the key design and contributions.
As shown in \tabref{table:main_result}, LiteSDMatte exhibits only a slight performance degradation compared to SDMatte, while still outperforming previous state-of-the-art methods.

\begin{table*}[htbp]
    \centering
    \renewcommand{\arraystretch}{1}
    \setlength{\tabcolsep}{2mm}{
    \resizebox{0.9\textwidth}{!}{
    \begin{tabular}{c|c|cc|cc|cc|cc|c}
        \toprule
        \multirow{2}{*}{\shortstack{Opacity\\Embedding}} & \multirow{2}{*}{\shortstack{Coordinate\\Embedding}} & \multicolumn{2}{c|}{AIM-500 (point)} & \multicolumn{2}{c|}{RefMatte-RW-100(point)} & \multicolumn{2}{c|}{AIM-500 (box)} & \multicolumn{2}{c|}{RefMatte-RW-100(box)} & \multirow{2}{*}{\shortstack{Impro $\uparrow$}} \\ 
         &  & MSE $\downarrow$ & SAD $\downarrow$ & MSE $\downarrow$ & SAD$\downarrow$ & MSE $\downarrow$ & SAD $\downarrow$ & MSE $\downarrow$ & SAD $\downarrow$  \\ 
         \midrule
          &  & 0.0169 & 44.23 & 0.0115 & 26.54 & 0.0098 & 28.55 & 0.0054 & 14.63 & - \\
          \checkmark &  & 0.0149 & 44.03 & 0.0111 & 26.65 & 0.0079 & 24.77 & 0.0060 & 15.52 & 3.85\% \\
          & \checkmark & 0.0167 & 45.17 & 0.0104 & 24.56 & 0.0109 & 28.85 & 0.0050 & 13.95 & 1.98\% \\
         \checkmark & \checkmark & 0.0139 & 40.18 & 0.0107 & 25.14 & 0.0077 & 24.26 & 0.0052 & 14.29 & \colorbox{first}{10.20\%} \\
        \bottomrule
    \end{tabular}}}
    \vspace{-5pt}
    \caption{\textbf{Ablation of Opacity Embedding and Coordinate Embedding.} Opacity embeddings represent the opacity information of objects, while coordinate embeddings encode the spatial position information from the visual prompts. The baseline is the setting that excludes opacity embedding and coordinate embedding.}
    \label{table:emb}
    \vspace{-5pt}
\end{table*}

\begin{table*}[htbp]
    \centering
    \renewcommand{\arraystretch}{1}
    \setlength{\tabcolsep}{2mm}{
    \resizebox{0.9\textwidth}{!}{
    \begin{tabular}{c|c|c|cc|cc|cc|cc|c}
        \toprule
        \multirow{2}{*}{\shortstack{Down\\Blocks}} & \multirow{2}{*}{\shortstack{Mid\\Block}} & \multirow{2}{*}{\shortstack{Up\\Blocks}} & \multicolumn{2}{c|}{AIM-500 (point)} & \multicolumn{2}{c|}{RefMatte-RW-100(point)} & \multicolumn{2}{c|}{AIM-500 (box)} & \multicolumn{2}{c|}{RefMatte-RW-100(box)} & \multirow{2}{*}{\shortstack{Impro $\uparrow$}} \\ 
         &  &  & MSE $\downarrow$ & SAD $\downarrow$ & MSE $\downarrow$ & SAD$\downarrow$ & MSE $\downarrow$ & SAD $\downarrow$ & MSE $\downarrow$ & SAD $\downarrow$  \\ 
         \midrule
          &  &  & 0.0101 & 30.62 & 0.0879 & 165.81 & 0.0075 & 23.78 & 0.0272 & 61.94 & - \\
         \checkmark &  &  & 0.0058 & 20.61 & 0.1378 & 253.08 & 0.0093 & 27.79 & 0.0227 & 51.12 & -5.12\% \\
           & \checkmark & & 0.0055 & 20.43 & 0.1360 & 245.48 & 0.0060 & 21.34 & 0.0368 & 84.25 & -8.12\% \\
           &  & \checkmark & 0.0074 & 24.04 & 0.0607 & 112.46 & 0.0052 & 20.07 & 0.0066 & 17.66 & 38.10\% \\
       \checkmark & \checkmark &  & 0.0055 & 20.99 & 0.1393 & 254.70 & 0.0077 & 24.49 & 0.0336 & 77.98 & -11.27\% \\
       \checkmark &  & \checkmark & 0.0128 & 35.56 & 0.0096 & 22.29 & 0.0073 & 23.10 & 0.0054 & 14.41 & 36.90\%\\
        & \checkmark & \checkmark & 0.0046 & 18.78 & 0.0714 & 134.23 & 0.0050 & 20.02 & 0.0052 & 13.86 & 42.32\% \\
        \checkmark & \checkmark & \checkmark & 0.0114 & 32.81 & 0.0099 & 22.54 & 0.0052 & 20.13 & 0.0060 & 14.60 & \colorbox{first}{44.43\%} \\
        \bottomrule
    \end{tabular}}}
    \vspace{-5pt}
    \caption{\textbf{Ablation of Masked Self-Attention Mechanism.} We apply the masked self-attention mechanism in various modules of the SDMatte to evaluate its sensitivity across different modules and identify the optimal performance setting. The setting without masked self-attention mechanism is considered the baseline.}
    \label{table:masked attn}
    \vspace{-5pt}
\end{table*}

\subsection{Ablation Studies}
In this section, we conduct a comprehensive set of ablation experiments to validate the effectiveness of our proposed design. 
All ablation experiments use the same training settings as the best result, except for the ablated parts.

\noindent \textbf{Visual Prompt-driven Cross-Attention Mechanism:} Diffusion models acquire strong text-driven interaction capabilities through training on large-scale data, enabling image generation conditioned on textual descriptions.
To leverage the powerful interaction capabilities of diffusion models and transfer them effectively to the interactive matting domain without disrupting the pre-trained weights, we propose a visual prompt-driven cross-attention mechanism.

We conduct ablation experiments to validate the effectiveness of this mechanism and evaluate its impact on performance across different blocks. 
As shown in \tabref{table:cross attn}, the results show that the visual prompt-driven cross-attention mechanism effectively inherits the text-driven interaction capability of the stable diffusion model.
Furthermore, experiments show that applying this mechanism solely to the middle block of the U-Net, where semantic information is most concentrated, leads to optimal performance, achieving an overall improvement of 11.67\% across two evaluation benchmarks and two types of visual prompts.

\noindent \textbf{Opacity Embedding and Coordinate Embedding:} In SDXL~\cite{podell2023sdxl}, image size and cropping parameters are incorporated as conditional inputs to the U-Net. This design enhances the model's robustness to diverse input sizes and produces centered outputs during inference. 
Inspired by this, we incorporate the coordinates of visual prompts and the opacity information of target objects into the U-Net, thereby improving the model's sensitivity to spatial position and opacity of objects.
Additionally, we adopt the one-step deterministic paradigm to accelerate inference speed and reduce the generation of erroneous details. Given that this paradigm does not require time embedding to represent the noise intensity, we empirically remove it.

To validate the effectiveness of our design, we conduct corresponding ablation experiments. As shown in \tabref{table:emb}, the opacity embeddings improve SDMatte's performance exclusively on the AIM benchmark, which contains numerous transparent foreground objects. In contrast, the coordinate embeddings of visual prompts enhance SDMatte's performance on the RefMatte-RW-100 benchmark, which serves as a multi-instance test set.
Additionally, the simultaneous use of coordinate embeddings and opacity embeddings results in a more comprehensive performance improvement of 10.20\% across two evaluation benchmarks, thereby validating the effectiveness of our design.

\noindent \textbf{Masked Self-Attention Mechanism:} 
To validate the effectiveness of the masked self-attention mechanism and its impact on performance across different blocks, we conduct corresponding ablation experiments.
As shown in \tabref{table:masked attn}, this mechanism contributes significantly to the down and up blocks of SDMatte. Its removal in either block impairs the module’s capacity to capture spatial location information, resulting in an emphasis on salient object extraction only.
Additionally, experimental results demonstrate that applying this mechanism to all modules of U-Net enables SDMatte to achieve both prediction accuracy and spatial awareness, leading to a more comprehensive improvement, which is regarded as the optimal configuration.

\section{Conclusion}
We propose SDMatte, an interactive matting method based on diffusion models.
This method effectively utilizes the rich prior knowledge of Stable Diffusion v2 and converts its text-driven interaction capability into a visual prompt-driven interaction capability through the visual prompt-driven cross-attention mechanism, leading to enhanced generalization and precise alpha matte predictions. 
By integrating coordinate and opacity embeddings, SDMatte achieves remarkable improvements in capturing spatial position information and object opacity information.
Additionally, we propose a masked self-attention mechanism to fully leverage the visual prompts, enabling the model to focus more on the regions indicated by visual prompts. Extensive experiments validate the effectiveness of our approach, which achieves state-of-the-art performance.
{
    \small
    \bibliographystyle{ieeenat_fullname}
    \bibliography{main}
}

\end{document}